\pdfoutput=1
\documentclass[11pt]{article}

\usepackage{EMNLP2023}

\usepackage{times}
\usepackage{latexsym}

\usepackage[T1]{fontenc}

\usepackage[utf8]{inputenc}
\usepackage{microtype}
\usepackage{hyperref}
\usepackage{url}
\usepackage{booktabs}

\usepackage{times}
\usepackage{latexsym}
\usepackage{spverbatim}
\usepackage{longtable}
\usepackage{multicol}
\usepackage{array}
\usepackage{makecell}
\usepackage{adjustbox}
\usepackage{xspace}
\usepackage{enumitem}
\usepackage{tabularx}
\usepackage{lineno}

\definecolor{darkblue}{rgb}{0, 0, 0.5}
\hypersetup{colorlinks=true, citecolor=darkblue, linkcolor=darkblue, urlcolor=darkblue}

\newcommand\system{\textsc{ROBoto2}\xspace}

\newcolumntype{L}[1]{>{\raggedright\let\newline\\\arraybackslash\hspace{0pt}}p{#1}}
\newcolumntype{M}[1]{>{\raggedright\let\newline\\\arraybackslash\hspace{0pt}}m{#1}}

\definecolor{darkgreen}{rgb}{0.0, 0.4, 0.13}
\definecolor{lightlightblue}{rgb}{0.9, 0.95, 1.0}

\title{\system: An Interactive System and Dataset for LLM-assisted \\Clinical Trial Risk of Bias Assessment}

\author{Anthony Hevia$^{1}$\Thanks{ denotes equal contribution} \quad Sanjana Chintalapati$^{1}$\footnotemark[1] \quad Veronica Ka Wai Lai$^2$ \\ \textbf{Thanh Tam Nguyen$^3$} \quad \textbf{Wai-Tat Wong$^4$} \quad \textbf{Terry Klassen$^5$} \quad \textbf{Lucy Lu Wang$^{1}$} \\ [0.5mm]
$^1$University of Washington \quad 
$^2$The Hospital for Sick Children \quad $^3$University of Bologna \\ $^4$The Chinese University of Hong Kong \quad $^5$University of Saskatchewan \\
\texttt{\{hevia, lucylw\}@uw.edu}
}

\begin{document}
\maketitle
\begin{abstract}
We present \system, an open-source, web-based platform for large language model (LLM)-assisted risk of bias (ROB) assessment of clinical trials. \system streamlines the traditionally labor-intensive ROB v2 (ROB2) annotation process via an interactive interface that combines PDF parsing, retrieval-augmented LLM prompting, and human-in-the-loop review. Users can upload clinical trial reports, receive preliminary answers and supporting evidence for ROB2 signaling questions, and provide real-time feedback or corrections to system suggestions. \system is publicly available at \href{https://roboto2.vercel.app/}{https://roboto2.vercel.app/}, with code and data released to foster reproducibility and adoption. We construct and release a dataset of 521 pediatric clinical trial reports (8954 signaling questions with 1202 evidence passages), annotated using both manually and LLM-assisted methods, serving as a benchmark and enabling future research. Using this dataset, we benchmark ROB2 performance for 4 LLMs and provide an analysis into current model capabilities and ongoing challenges in automating this critical aspect of systematic review.\footnote{Dataset and code at \href{https://github.com/larchlab/ROBoto2}{https://github.com/larchlab/ROBoto2}}
\end{abstract}

\section{Introduction}
Clinical trials, especially when aggregated in systematic reviews, provide the highest quality of evidence for clinical care. 
While many steps in the systematic review pipeline have seen increasing automation \citep{Marshall2019TowardSR,Khalil2021ToolsTS,Alshami2023HarnessingTP}, especially with the advent of LLMs and associated technology, assessing the quality of evidence in individual trials, specifically evaluating \emph{risk of bias} (ROB), remains a critical and time-consuming bottleneck.

The Cochrane Risk of Bias tool version 2 (ROB2)\footnote{\href{https://methods.cochrane.org/bias/resources/rob-2-revised-cochrane-risk-bias-tool-randomized-trials}{https://methods.cochrane.org/bias/resources/rob-2-revised-cochrane-risk-bias-tool-randomized-trials}} standardizes evaluation by asking 22 signaling questions over 5 domains and computing an overall judgment about risk of bias.
However, applying ROB2 is time-consuming, taking trained reviewers 30+ minutes per clinical trial report. This limits scalability for large systematic reviews synthesizing hundreds or thousands of trials.

\begin{figure*}[t!]
    \centering
    \includegraphics[width=\textwidth]{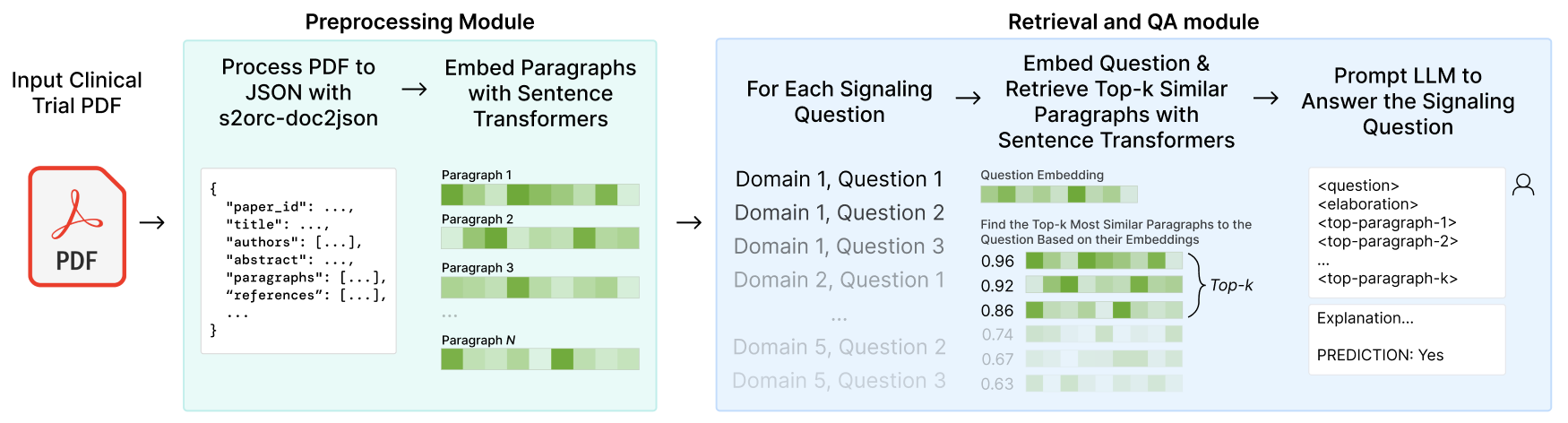}
    \caption{\system system pipeline. Given a clinical trial PDF as input, \system first preprocesses the document to extract and embed paragraphs. Then, a QA module iterates through all of the questions of the ROB2 assessment to identify evidence passages and prompt GPT3.5 to answer the question based on the retrieved evidence.}
    \label{fig:pipeline}
\end{figure*}

Previous systems such as RobotReviewer \citep{marshall2016robotreviewer} and others \citep{Marshall2014AutomatingRO} explored automating an earlier version of the ROB assessment (ROB) via supervised models, but practical, high-quality automation for ROB2 remains elusive.
We therefore introduce \system, a web-based platform supporting human-AI collaborative ROB2 assessment. 
\system integrates PDF parsing, within-document evidence retrieval, LLM prompting, and ROB2 logic to provide initial answers and rationales for each signaling question. Experts can accept, modify, or override suggestions, with feedback captured for future improvement. Using \system, our medical collaborators conducted ROB2 assessments on 521 pediatric clinical trials---245 via fully manual review and 276 using the LLM-assisted workflow---yielding a new dataset for benchmarking and research.

We evaluate retrieval methods and four LLMs (Llama-3.3-70B-Instruct, GPT-3.5-Turbo, GPT-4o, and Claude 3.5-Sonnet) on the 245 manual assessments subset, finding that LLMs remain overly conservative compared to human reviewers, frequently opting for high-risk or ``No Information'' judgments even when evidence is present. Larger context windows and more retrieved evidence somewhat mitigate these tendencies, but fully automated, accurate ROB2 assessment remains challenging.

To summarize, we contribute the following:
\begin{itemize}[noitemsep, topsep=0pt, leftmargin=10pt]
     \item We introduce the \system system, a public web tool (code and API available) that supports a human-AI collaborative pipeline for clinical trial ROB2 assessment; the system integrates document preprocessing, passage retrieval, LLM prompting, and interactive expert review;
    \item We release a dataset of 521 ROB2 assessments (8954 questions; 1202 evidence passages), including both manual and LLM-assisted annotations by medical experts, conducted in the context of an ongoing, real-world systematic review of pediatric clinical trial literature;
    \item We benchmark retrieval strategies and 4 LLMs on this dataset, providing the first evaluation of LLM-assisted ROB2 assessment. Our analysis highlights current model limitations and directions for future improvement.
\end{itemize}

\section{Related Work}
\label{sec:rw}

\paragraph{Automating systematic review} Prior work on automating systematic reviews have investigated ways to automate the retrieval of relevant papers on a review topic \citep{Choong2014AutomaticER, Portenoy2020ConstructingAE, Schoot2021ASReviewAL}, gauging the quality of clinical trials via risk of bias assessment \citep{Marshall2014AutomatingRO, marshall2016robotreviewer, Suster2021AutomatingQA}, extracting PICO (population, intervention, comparator, outcome) elements \citep{Wallace2016ExtractingPS, nye-etal-2018-corpus, Jin2018PICOED, Hu2023TowardsPP}, extracting numerical results \citep{Yun2024AutomaticallyEN, naik-etal-2024-care}, classifying the direction of evidence, also called evidence inference \citep{lehman-etal-2019-inferring, deyoung-etal-2020-evidence}, as well as synthesizing and summarizing results across different studies \citep{Wallace2020GeneratingN, deyoung-etal-2021-ms, wang-etal-2022-overview, SanchezGraillet2022SynthesizingEF, shaib-etal-2023-summarizing}. Our work builds upon this prior work, especially towards assessing the quality of trials via LLM-assisted risk of bias analysis, extending to v2 of the ROB tool. 

\paragraph{ROB analysis} The ROB assessment questionnaire from~\citet{higgins2011cochrane} and~\citet{Sterne2019RoB2A} can be used to determine the extent to which randomized control trials are at risk of bias. \citet{Suster2021AutomatingQA} provide quality ratings for bodies of evidence, and found that some risk factors for quality have good accuracy when automatically assessed, while others do not due to data scarcity. RobotReviewer~\citep{marshall2016robotreviewer} introduced a system that automatically assigns ROB categorizations to randomized control trials using a trained language model. We extend this work by (i) introducing a dataset corresponding to the newer and more reliable version of the ROB tool (ROB2) \citep{Sterne2019RoB2A}, (ii) creating an annotation system geared towards supporting a researcher in the loop~\cite{Jardim2022AutomatingRO}, which leverages in-document retrieval and LLMs to answer signaling questions and identify rationales from the source articles, and (iii) conducting experiments and analysis demonstrating the performance and limitations of current LLMs in supporting this task. 

\section{Background}
\label{sec:background}

We measure risk of bias of randomized trials using the Cochrane ROB2 tool.\footnote{ROB2 replaces its predecessor ROB after a formal evaluation
identified areas for improvement \citep{Sterne2019RoB2A}. ROB2 includes questions measuring newly identified ways that bias arise in randomized trials.}
The ROB2 tool assesses risk along five domains that can introduce bias into the results of a randomized trial:

\begin{itemize}[noitemsep, topsep=0pt, itemindent=6pt]
  \item[D1:] Randomization process
  \item[D2:] Deviations from intended interventions
  \item[D3:] Missing outcome data
  \item[D4:] Measurement of the outcome
  \item[D5:] Selection of the reported result
\end{itemize}

\noindent Each domain consists of 3-7 signaling questions, which help gather information and contribute to the final risk classification. For example, this D2 question assesses bias due to unblinded treatment assignment:~``Were participants aware of their assigned intervention during the trial?'' There are five response options for each signaling question: (1) Yes; (2) Probably yes; (3) Probably no; (4) No; and (5) No information. All questions in App.~\ref{app:rob2}.

The ROB2 assessment is hierarchical. Signaling question responses for each domain first contribute to domain-level judgments for risk of bias, then domain-level judgments provide the basis for an overall risk of bias judgment. The tool provides flowcharts for computing the risk of each domain based on the answers to signaling questions (e.g., Figure~\ref{fig:flowchart} in App.~\ref{app:rob2}) as well as for computing overall risk. Domain-level and overall risk are assessed as either low risk, some concerns, or high risk.

In \system, we model the ROB2 assessment as a document-level question-answering (QA) task. We use each signaling question as a query to retrieve relevant evidence passages from the trial report, then generate an answer based on the retrieved evidence. Answers are validated by a user who is conducting the assessment. The final risk assessment is produced by implementing the flowchart logic provided by the ROB2 tool.

\section{\system System Pipeline}
\label{sec:pipeline}

\begin{figure*}[t!]
    \centering
    \includegraphics[width=\textwidth]{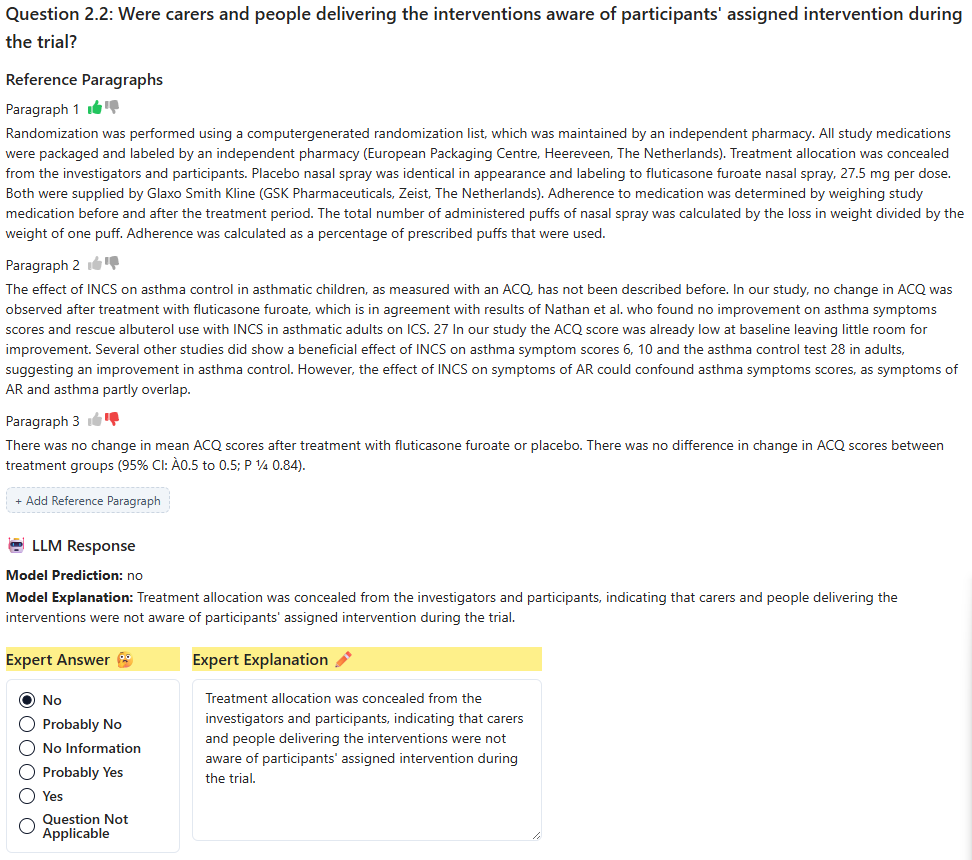}
    \caption{Screenshot of \system assisting with a question from Domain 2. The user can modify the model-provided answer and explanation and rate reference paragraphs.}
    \label{fig:rob2_sys_example}
\end{figure*}

Figure~\ref{fig:pipeline} shows the \system pipeline. A user uploads a PDF of a clinical trial report. We (i) preprocess it to extract paragraphs of text; (ii) embed each paragraph using a document embedding model and index them for within-document retrieval; and then, for each signaling question from the ROB2 assessment tool, we (iii) embed the signaling question, retrieve the top-$k$ similar paragraphs from the paper, and prompt an LLM to answer the question using the top-$k$ paragraphs as context. We also experiment with providing all passages of text (full paper) as input for models with large input context windows. 
Details follow. 

\vspace{-1mm}
\paragraph{Preprocessing PDFs}
We convert each PDF into standardized JSON format using the S2ORC-doc2json library~\citep{Lo2020S2ORCTS}.\footnote{\href{https://github.com/allenai/s2orc-doc2json}{https://github.com/allenai/s2orc-doc2json}} The output JSON contains a list of paragraphs in the paper, their section headers, and metadata elements such as the paper's title, authors, and abstract.

\vspace{-1mm}
\paragraph{Embedding paragraphs for retrieval}
We compute embeddings for each paragraph in the uploaded paper using Sentence-Transformers all-MiniLM-L6-v2 \citep{reimers-2019-sentence-bert} and construct a key-value store for retrieval. Evaluation of the retriever and alternate methods is described in App.~\ref{app:retriever-eval}.
For each signaling question, we embed the question text using the same model and use cosine similarity to identify the top-$k$ paragraphs to use as context for the QA module. 

\vspace{-1mm}
\paragraph{Answering signaling questions}
We then prompt an LLM to answer each ROB2 signaling question using an instruction prompt based on the ROB2 questionnaire and with the retrieved evidence paragraphs as context. Prompt templates and a complete example are given in App.~\ref{app:prompt}. 

\vspace{-1mm}
\paragraph{Collecting user feedback} When conducting assessments with \system, users can modify and provide feedback on all aspects of the assessment. While we show the top-3 retrieved paragraphs by default, users can add further paragraphs by selecting from the JSON parse. They can also provide feedback on the accuracy of retrieved passages via up- or downvotes, and modify the model-predicted answers and rationales (called ``Explanation'' in \system). \system retains the original LLM responses and rationales, as well as the versions confirmed or edited by expert users. We include these user modifications as part of our dataset. 

\vspace{-1mm}
\paragraph{Domain-level and overall judgments} We implement the logic provided by the ROB2 flowcharts (e.g., Figure~\ref{fig:flowchart} in App.~\ref{app:rob2}) to produce domain-level and overall risk of bias judgments. We visualize these at the end of each domain section and as a summative visualization when users complete the ROB2 assessment, e.g., three high risk domain-level judgments yields a ``High Risk'' overall rating visualized as follows: 
\vspace{-3mm}
\begin{center}
\noindent
\colorbox{gray!8}{%
  \makebox[0.98\linewidth][c]{
    \includegraphics[width=0.48\linewidth]{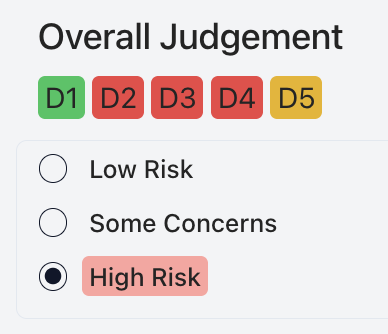}%
  }%
}
\end{center}
\vspace{-1mm}
\paragraph{System implementation} Part of the \system workflow is shown in Figure~\ref{fig:rob2_sys_example} (web app at \href{https://roboto2.vercel.app/}{https://roboto2.vercel.app/}). The web interface is written in React and Typescript. It leverages Transformers.js for client-side embedding and retrieval, and a back-end API in Python and FastAPI for document parsing and calls to LLM services.

\section{Annotated Evaluation Dataset}
\label{sec:dataset}

Our dataset consists of 521 ROB2 assessments (245 completed using the Cochrance ROB2 Excel tool and 276 with LLM assistance via \system). These ROB2 assessments were conducted as part of an independent research project aiming to systematically evaluate the risk of bias of all child health clinical trials; we re-purpose the data in this work to explore the role and feasibility of LLMs in supporting this aspect of systematic review.

\vspace{-1mm}
\paragraph{Annotation procedure}
An initial corpus of child health clinical trial reports was constructed by searching the Cochrane Central Register of Controlled Trials, filtering for pediatric clinical trials based on the procedures described in \citet{Boluyt2008UsefulnessOS}, and identifying 2334 matching clinical trial reports published 1991-2020. We sampled trial reports from this corpus for annotation. 

For a subset of 245 reports, a group of expert raters completed assessments manually using the Cochrane tool, an Excel sheet with macros implementing the logic of the ROB2 assessment. To support judgments, annotators identified evidence passages manually from paper PDFs for a subset of questions and copied these into the Excel sheet. For each clinical trial, the data consists of the paper PDF for the trial report, as well as judgments for each signaling question, evidence passages extracted from the paper for a subset of questions, domain-level judgments, and the overall risk assessment score. Five expert annotators participated in annotations, and all annotators have graduate degrees in public health, epidemiology, medical sciences, or clinical practice, as well as experience conducting systematic reviews. This set of 245 papers, annotated using the current gold-standard ROB2 review process, is used for all LLM evaluation and comparison reported in this paper. 

An additional 276 papers were annotated separately by two of the five annotators using \system, with LLM assistance. The version of \system used for annotations (collected during early 2024) used retrieval and GPT-3.5 (gpt-3.5-turbo-0125) as the answer model.\footnote{In experiments, other LLMs and full-text context demonstrate better performance, but these were reasonable configurations at the time of annotation. Our public web interface supports the use of alternate LLMs.} Because this sample of 276 papers was annotated with LLM assistance, we withheld them from the final evaluation of \system as described in Section~\ref{sec:experiments}, but include them in the published dataset to support future work. 

\vspace{-1mm}
\paragraph{Dataset statistics}
The distribution of domain-level and overall risk of bias judgments in the full dataset are provided in Table~\ref{tab:data_distribution}. Distributions of answered signaling questions and evidence paragraphs in the manually-annotated subset are shown at the top of Table~\ref{tab:results}.

\begin{table}[t!]
\small
\centering
    \begin{tabular}{lccc}
        \toprule
         & Low risk & Some concerns & High risk \\
        \midrule
        Domain 1 & 234 & 243 & 44 \\
        Domain 2 & 287 & 171 & 63 \\
        Domain 3 & 450 & 35 & 35 \\
        Domain 4 & 406 & 60 & 54 \\
        Domain 5 & 332 & 272 & 34 \\
        \midrule
        Paper-level & 64 & 301 & 156 \\
        \bottomrule
    \end{tabular}
    \caption{Distribution of domain- and paper-level risk of bias judgments in our dataset.}
    \label{tab:data_distribution}
\end{table}

\vspace{-1mm}
\paragraph{Inter-rater reliability} 
To assess inter-rater reliability, 20 papers (totaling 440 signaling questions) are independently annotated by two annotators using \system. We aggregate answers into the following classes: Yes/Probably Yes, No/Probably No, No Information, and N/A, when a question is skipped by ROB logic. Four-class Cohen's Kappa is 0.40, indicating fair to moderate agreement.
This is consistent with prior work showing slight to moderate agreement (Fleiss’ Kappa of 0.45 at the domain level) among experienced raters \citep{Minozzi2020TheRC, Minozzi2021ReliabilityOT}; it reflects well-documented challenges of applying the complex ROB2 tool \citep{Nejadghaderi2024TheCR}, as reviewers can differ in their interpretation of ambiguous scenarios and their thresholds for assigning risk levels. Further commentary in App.~\ref{app:inter-rater-reliability-analysis}.

\section{Experimental Settings}
\label{sec:experiments}
\vspace{-1mm}
\begin{table*}[th!]
    \centering
    \small
    \begin{tabular}{L{36mm}llcccccc}
        \toprule
         Model & Retrieval & D1 & D2 & D3 & D4 & D5 & Micro-Avg & Macro-Avg \\
        \midrule
        $n$-oracle & - & 197 & 124 & 37 & 73 & 11 & - & - \\
        $n$-total & - & 750 & 1278 & 598 & 1027 & 750 & - & - \\
        \midrule
        \multicolumn{9}{l}{\textbf{Baseline w/ oracle evidence paragraphs}} \\[0.5mm]
        Llama-3.3-70B-Instruct & Oracle & 0.67 & 0.55 & 0.35 & 0.81 & 0.45 & 0.62 & 0.67 \\
        GPT-3.5-Turbo & Oracle & 0.81 & 0.66 & 0.67 & 0.82 & 0.57 & 0.61 & 0.71 \\
        GPT-4o & Oracle & 0.75 & 0.78 & 0.68 & 0.92 & 0.54 & 0.64 & 0.73 \\
        Claude 3.5-Sonnet & Oracle & 0.75 & 0.81 & 0.66 & 0.92 & 0.59 & 0.66 & 0.75 \\
        \midrule
        \multicolumn{9}{l}{\textbf{Retrieved evidence paragraphs}} \\[0.5mm]
        Llama-3.3-70B-Instruct & k=1 & 0.83 & 0.61 & 0.42 & 0.80 & 0.38 & 0.49 & 0.61 \\
        GPT-3.5-Turbo & k=1 & 0.81 & 0.73 & 0.69 & 0.74 & 0.66 & 0.58 & 0.73 \\
        GPT-4o & k=1 & 0.68 & 0.65 & 0.58 & 0.81 & 0.75 & 0.55 & 0.69 \\
        Claude 3.5-Sonnet & k=1 & 0.69 & 0.68 & 0.66 & 0.87 & 0.76 & 0.60 & 0.73 \\
        \midrule
        
        Llama-3.3-70B-Instruct & k=3 & 0.87 & 0.69 & 0.66 & 0.81 & 0.35 & 0.55 & 0.68 \\
        GPT-3.5-Turbo & k=3 & 0.82 & 0.69 & 0.68 & 0.73 & 0.59 & 0.55 & 0.70 \\
        GPT-4o & k=3 & 0.75 & 0.72 & 0.73 & 0.82 & 0.72 & 0.60 & 0.75 \\
        Claude 3.5-Sonnet & k=3 & 0.75 & 0.75 & 0.76 & 0.87 & \textbf{0.77} & 0.65 & 0.78 \\
        \midrule
        
        Llama-3.3-70B-Instruct & k=5 & 0.87 & 0.72 & 0.70 & 0.81 & 0.28 & 0.55 & 0.69 \\
        GPT-3.5-Turbo & k=5 & \textbf{0.82} & 0.69 & 0.64 & 0.71 & 0.56 & 0.53 & 0.68 \\
        GPT-4o & k=5 & 0.78 & 0.74 & 0.73 & 0.83 & 0.69 & 0.62 & 0.75 \\
        Claude 3.5-Sonnet & k=5 & 0.78 & 0.79 & 0.78 & 0.86 & \textbf{0.77} & 0.67 & 0.80 \\
        \midrule

        \multicolumn{9}{l}{\textbf{Full paper as input}} \\[0.5mm]
        Llama-3.3-70B-Instruct & Full Paper & 0.88 & 0.81 & 0.79 & 0.79 & 0.32 & 0.61 & 0.72 \\
        GPT-4o & Full Paper & 0.80 & 0.82 & 0.77 & 0.85 & 0.66 & 0.66 & 0.78 \\
        Claude 3.5-Sonnet & Full Paper & 0.81 & \textbf{0.84} & \textbf{0.80} & \textbf{0.88} & \textbf{0.77} & \textbf{0.71} & \textbf{0.82} \\
        \bottomrule
    \end{tabular}
    \vspace{-1mm}
    \caption{For all settings, we report micro-averaged domain-level F1 along with micro- and macro-averaged F1 across all signaling questions. The $n$-oracle is the number of instances where annotators identified an evidence passage, while $n$-total is the total number of signaling questions answered for that domain.}
    \label{tab:results}
    \vspace{-1mm}
\end{table*}

We evaluate 4 models: GPT-3.5-Turbo, GPT-4o, Claude 3.5-Sonnet, and Llama-3.3-70B-Instruct. All models receive the same prompt and inputs (App.~\ref{app:prompt}). For all experiments, we aggregate labels and outputs into three classes: Yes/Probably Yes (Y/PY), No/Probably No (N/PN), and No Information (NI), and report micro-F1 at each domain level along with micro- and macro-averages across all signaling questions (Table~\ref{tab:results}).

\vspace{-1mm}
\paragraph{Within-document retrieval} 
We evaluate two retrieval methods: BM25 \citep{Robertson1994OkapiAT} and paragraph embeddings using Sentence-Transformers \citep{reimers-2019-sentence-bert}. Each signaling question has a max of one gold evidence passage in the dataset; we report recall@$k$ for $k$=1,3,5,10 for all retrieval methods (Table~\ref{tab:retriever-results-table}). Detailed results and evaluation of the retrieval methods can be found in App~\ref{app:retriever-eval}. In all cases, we use the questions from the ROB2 assessment as the query, and paragraphs from the clinical trial paper as the documents to retrieve. In the publicly available version of \system, all-MiniLM-L6-v2 and $k$=3 were selected as these settings achieved competitive performance at low cost. 

\vspace{-1mm}
\paragraph{Prompting LLMs for QA} 
We evaluate all models in a zero-shot setting with oracle evidence (providing the human-labeled evidence passage), as well as with the top-$k$ retrieved evidence (with k=1,3,5), and the full paper setting for models with sufficient input context window sizes (all but GPT-3.5-Turbo).
In the ROB2 assessment, each signaling question includes elaboration text that expands on when each answer should be chosen for that question; we provide this elaboration in the instructions for all prompting settings (example in App.~\ref{app:prompt}). We also conduct several experiments with in-context learning~\citep{brown2020language}, which suggested minimal gains from the zero-shot setting; these results are reported in App~\ref{app:fs-prompting-results}.

\section{Results \& Discussion}
\label{sec:discussion}
\vspace{-1mm}
Results for all experimental settings are provided in Table~\ref{tab:results}.  
We analyze the model results and user statistics collected during \system annotations below (full statistics in App.~\ref{app:annotator-para-pref}).

\vspace{-1mm}
\paragraph{Room for improvement}
The best performing model (Claude 3.5-Sonnet with the full paper as context) achieved a micro-F1 of 0.71, highlighting considerable room for improvement. All evaluated models achieve strong results in D1, where questions are more likely to be answerable based on text in the trial reports. Performance in D2 and D3 is mixed, as these may require interpreting numerical data (e.g., calculating attrition rates from recruitment and result numbers). D5 is similarly challenging for models as these questions may require knowledge of external clinical resources and guidelines.

Increasing context generally improves performance for most models while reducing accuracy for GPT-3.5-Turbo. In some cases, models with retrieval can surpass oracle retrieval performance, likely due to incomplete evidence labeling in our dataset, indicating that relevant information exists beyond annotator-selected passages. Few-shot prompting does not appear to outperform zero-shot prompting in our experimental results (App~\ref{app:fs-prompting-results}).

\vspace{-1mm}
\paragraph{Limited utility for fully-automated ROB assessment}
Model performance cannot be substituted for human judgment in ROB assessments, and we recommend that humans remain in the loop. Conservative question-level model judgments compound to conservative domain- and paper-level judgments, where the fully-automated pipeline judges most papers as having ``some concerns'' or ``high risk'' even when human raters did not. Human raters assessed 47 of 276 trials as high risk while the LLM-only pipeline assessed 101 as high risk. Error analysis (App.~\ref{app:llm-error-analysis}) reveals that the strongest models tend to over-select ``No Information,'' which may reflect cautiousness gained from safety and alignment training. 

\vspace{-1mm}
\paragraph{\system supports human review and editing} We compute detailed metrics for the 276 ROB2 assessments annotated using \system, including the number of times annotators accept the model's answers and explanations directly versus change them, and the number of retrieved evidence passages marked as good (offering evidence to support an answer) versus bad (irrelevant). 
Annotators provide their own answer (42.4\%) and edit rationales (28.7\%) around half the time, rather than use the answer (57.6\%) and rationales (71.3\%) provided by the model (Table~\ref{tab:usage-results-table}). For evidence passages, 615 total up/downvotes are collected (out of 3370 retrieved passages), of which 78.0\% are positive feedback.
We provide all feedback in our dataset to support future model development.
Detailed statistics can be found in App.~\ref{app:annotator-para-pref}.

\section{Conclusion}
\vspace{-1mm}
Assessing the quality of clinical trials is an important step to weighing their evidence in clinical decision-making. To support this, we introduce the \system system to assist researchers in conducting risk of bias assessment for clinical trials with LLM support, along with associated code for running the web interface. We also release a dataset of 521 complete ROB analyses (8954 signaling questions with 1202 evidence passages) of child health clinical trial reports.
We hope our system and dataset will promote better LLM applications for risk of bias assessment, and that access to this assisted annotation tool can enable quicker completion of ROB assessments and reduce the labor and costs around systematic literature reviews.

\section{Limitations}

\vspace{-1mm}
\paragraph{Viewing model outputs could potentially bias annotations}
\system is designed to expose all intermediate and final model outputs, and
allows expert annotators to change any part of the model output. While we can compute the number of changes made, we cannot guarantee that seeing model outputs does not influence annotator responses. Prior work has shown that human annotators may demonstrate anchoring bias when exposed to LLM assistance during annotation \citep{choi-etal-2024-llm}, leading to discrepancies in downstream label distributions. We leave the measurement of this bias in the ROB setting to future work. 

\vspace{-1mm}
\paragraph{Dataset imbalance}
Though it reflects real-world ROB2 assessments, our dataset is unbalanced. The majority of papers are assessed as having some concerns, with fewer papers of low or high risk. This likely biases the evaluation of our system, similar to what was observed by \citet{Suster2021AutomatingQA}. Related, some signaling questions have very sparse annotations (especially those that depend on cascading logic) or are biased in terms of answer distribution (almost always one of the answer labels). 

Among manually conducted reviews, annotator-provided evidence paragraphs are only available for a small portion of signaling questions, unbalanced across domains; D1 has the most signaling questions with evidence passages, while D5 has very few. Our retrieval methods as well as Oracle results are only reported on this biased subset, and may not accurately represent performance on sparsely annotated questions and domains.

\vspace{-1mm} 
\paragraph{Potential gains in quality or efficiency} We hypothesize that \system may either help to save time or improve the quality of ROB assessments. Qualitatively, the annotation team reported that \system offers an opportunity to enhance evidence and rationale coverage, as the time savings on the ROB assessment itself were repurposed to judge and retrieve relevant evidence. However, the impact of such repurposing of effort on the quality of the resulting assessments was not explicitly measured in our system and should be confirmed and studied in future work.

\vspace{-1mm}
\paragraph{Diversity of language models in experiments}
Our experiments do not include any reasoning models such as OpenAI's o3, Anthropic's Claude 4-Sonnet-Thinking, and DeepSeek-R1~\citep{deepseekai2025deepseekr1incentivizingreasoningcapability}. Future work could explore whether these models improve current performance in terms of both answer classification accuracy and generated rationales.

\section*{Acknowledgements}
We thank the following individuals for producing parts of the ROB dataset we use for evaluation: Bernadette Zakher, Shannon Sim, Michele Dyson, Banke Oketola, and Aneet Saran, from the University of Victoria, University of Alberta, and University of Manitoba. This work is supported in part by the University of Washington Information School Strategic Research Fund and the University of Washington eScience Institute's Azure Cloud Credits for Research and Teaching.

\bibliography{anthology, custom}
\bibliographystyle{acl_natbib}

\appendix

\section{ROB2 Assessment Tool}
\label{app:rob2}

The Signaling Questions for the Cochrane ROB2 Tool for Randomized Trials are given in Table~\ref{tab:rob2-questions}. Note that some questions are \emph{cascading}, and are only answered if previous questions in the domain are answered in a pre-specified way.

\begin{figure*}[th!]
    \centering
    \includegraphics[width=\textwidth]{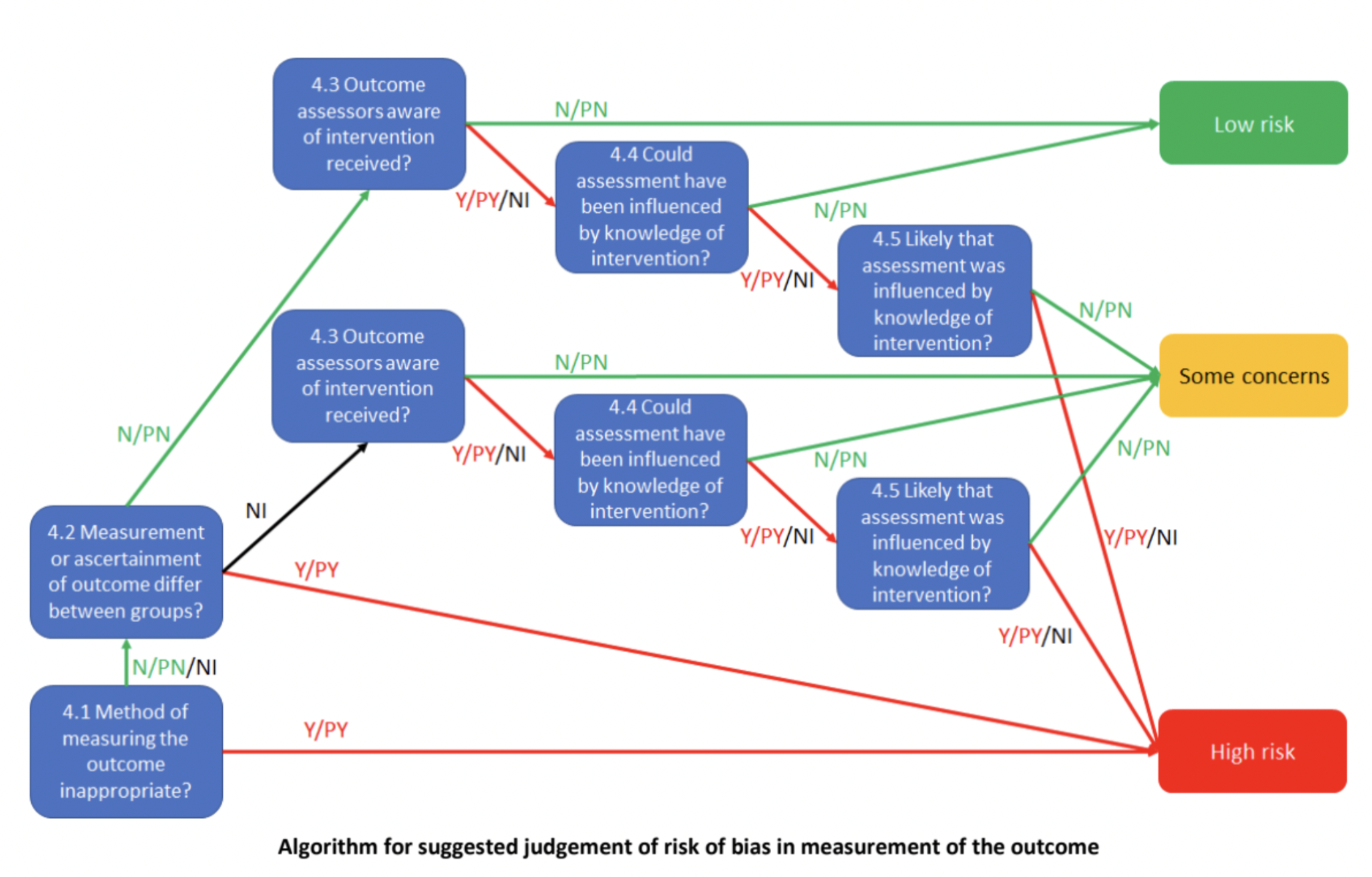}
    \caption{Flowchart for how answers to signaling questions contribute to a domain-level judgment for Domain 4 in the ROB2 tool. Reproduced from \href{https://sites.google.com/site/riskofbiastool/welcome/rob-2-0-tool}{https://sites.google.com/site/riskofbiastool/welcome/rob-2-0-tool}.}
    \label{fig:flowchart}
\end{figure*}

Figure~\ref{fig:flowchart} reproduces a flowchart from the ROB2 tool that indicates how signaling questions contribute to a domain-level judgment for Domain 4. Based on how these questions are answered, the domain-level judgment can be low risk, some concerns, or high risk. Flowcharts are also provided for the other four domains and are available at \href{https://methods.cochrane.org/risk-bias-2}{https://methods.cochrane.org/risk-bias-2}.

\begin{table*}[t!]
\centering
\small
\begin{tabular}{p{10mm}p{125mm}}
\toprule
\textbf{Domain} & \textbf{Question} \\
\midrule
\multicolumn{2}{l}{\textbf{Domain 1: Risk of bias arising from the randomization process}} \\ 
\midrule
1.1 & Was the allocation sequence random? \\
1.2 & Was the allocation sequence concealed until participants were enrolled and assigned to interventions? \\
1.3 & Did baseline differences between intervention groups suggest a problem with the randomization process? \\
\midrule
\multicolumn{2}{l}{\textbf{Domain 2: Risk of bias due to deviations from the intended interventions}} \\
\midrule
2.1 & Were participants aware of their assigned intervention during the trial? \\
2.2 & Were carers and people delivering the interventions aware of participants' assigned intervention during the trial? \\
2.3 & If Y/PY/NI to 2.1 or 2.2: Were there deviations from the intended intervention that arose because of the trial context? \\
2.4 & If Y/PY to 2.3: Were these deviations likely to have affected the outcome? \\
2.5 & If Y/PY/NI to 2.4: Were these deviations from intended intervention balanced between groups? \\
2.6 & Was an appropriate analysis used to estimate the effect of assignment to intervention? \\
2.7 & If N/PN/NI to 2.6: Was there potential for a substantial impact (on the result) of the failure to analyse participants in the group to which they were randomized? \\
\midrule
\multicolumn{2}{l}{\textbf{Domain 3: Risk of bias due to missing outcome data}} \\
\midrule
3.1 & Were data for this outcome available for all or nearly all participants randomized? \\
3.2 & If N/PN/NI to 3.1: Is there evidence that the result was not biased by missing outcome data? \\
3.3 & If N/PN to 3.2: Could missingness in the outcome depend on its true value? \\
3.4 & If Y/PY/NI to 3.3: Is it likely that missingness in the outcome depended on its true value? \\
\midrule
\multicolumn{2}{l}{\textbf{Domain 4: Risk of bias in measurement of the outcome}} \\
\midrule
4.1 & Was the method of measuring the outcome inappropriate? \\
4.2 & Could measurement or ascertainment of the outcome have differed between intervention groups? \\
4.3 & If N/PN/NI to 4.1 and 4.2: Were outcome assessors aware of the intervention received by study participants? \\
4.4 & If Y/PY/NI to 4.3: Could assessment of the outcome have been influenced by knowledge of intervention received? \\
4.5 & If Y/PY/NI to 4.4: Is it likely that assessment of the outcome was influenced by knowledge of intervention received? \\
\midrule
\multicolumn{2}{l}{\textbf{Domain 5: Risk of bias in selection of the reported result}} \\
\midrule
5.1 & Were the data that produced this result analysed in accordance with a pre-specified analysis plan that was finalized before unblinded outcome data were available for analysis? \\
5.2 & Is the numerical result being assessed likely to have been selected, on the basis of the results, from multiple eligible outcome measurements (e.g., scales, definitions, time points) within the outcome domain? \\
5.3 & Is the numerical result being assessed likely to have been selected, on the basis of the results, from multiple eligible analyses of the data? \\
\bottomrule
\end{tabular}
\caption{Signaling Questions in the Cochrane Risk of Bias Tool for Randomized Trials (ROB2).}
\label{tab:rob2-questions}
\end{table*}

\section{Retriever Evaluation}
\label{app:retriever-eval}
\begin{table}[t!]
    \centering
    \small
    \begin{tabular}{p{12mm} c c c c c}
        \toprule
        Model & R@1 & R@3 & R@5 & R@10\\
        \midrule\\[-0.15in]
        BM25 & 0.140 & 0.272 & 0.367 & 0.533 \\
        S-BERT & 0.268 & 0.455 & 0.519 & 0.678 \\
        \bottomrule
    \end{tabular}
    \caption{Recall@$k$ for our tested retrieval methods.}
    \label{tab:retriever-results-table}
\end{table}

We experiment with a sparse retriever, BM25 \citep{Robertson1994OkapiAT}, and Sentence-Transformers \citep{reimers-2019-sentence-bert}. We assess retriever performance by varying $k$, the number of passages retrieved and provided to the QA reader module. For models with large context windows, we also experiment with providing the entire paper as context.

All methods are validated using gold evidence paragraphs identified by the annotators in our dataset. 
Each signaling question has a maximum of one gold evidence passage in the dataset; we report recall@$k$ for $k$=1,3,5,10 for all retrieval methods (Table~\ref{tab:retriever-results-table}).

For within-document retrieval, we find that S-BERT successfully retrieves the gold evidence passage at a higher rate than BM25 at comparable $k$ (Table~\ref{tab:retriever-results-table}). However, prompting models with the full paper achieves the highest overall F1-scores (Table~\ref{tab:results}). \system uses S-BERT for its balance between performance, speed, and enabling models with smaller context windows to be used in the web interface.

\section{Prompting}
\label{app:prompt}

The prompt is formatted as follows: \\ [-2mm]

\footnotesize
\colorbox{lightlightblue}{\parbox{0.88\linewidth}{
\texttt{<instruction> \\
<signaling\_question> \\
<elaboration> \\
<retrieved\_paragraph\_1> \\
... \\
<retrieved\_paragraph\_k>
}}} \\

\normalsize

\noindent After an instruction to answer the question, we provide the signaling question itself from the ROB2 assessment, as well as additional elaboration text explaining all answer options and when they should be used. These elaborations are adapted from explanations given in the ROB2 tool, and we further augment them such that all possible answer options are represented---not all answers are represented in elaborations from the original ROB2 tool, which we found may bias models towards answers that were. Retrieved context paragraphs are then appended. We instruct the model to make a prediction and generate a rationale for its prediction. 

A full example prompt for signaling question 1 in Domain 1 is reproduced below:

\footnotesize
\begin{spverbatim}
You are an expert scientific researcher. You will be given a passage from a scientific paper reporting on a randomized controlled trial along with a question and elaboration of the question. Your task is to return the answer to the question out of the following set of answers: "yes", "no", "probably yes", "probably no", "no information". You should use the given passage to answer the question.

Question: "Was the allocation sequence random?"

Elaboration: "Answer ‘Yes’ if a random component was used in the sequence generation process. Examples include computer-generated random numbers; reference to a random number table; coin tossing; shuffling cards or envelopes; throwing dice; or drawing lots. Minimization is generally implemented with a random element (at least when the scores are equal), so an allocation sequence that is generated using minimization should generally be considered to be random.

Answer ‘No’ if no random element was used in generating the allocation sequence or the sequence is predictable. Examples include alternation; methods based on dates (of birth or admission); patient record numbers; allocation decisions made by clinicians or participants; allocation based on the availability of the intervention; or any other systematic or haphazard method.

Answer ‘No information’ if the only information about randomization methods is a statement that the study is randomized.

In some situations a judgment may be made to answer ‘Probably no’ or ‘Probably yes’. For example, in the context of a large trial run by an experienced clinical trials unit, absence of specific information about generation of the randomization sequence, in a paper published in a journal with rigorously enforced word count limits, is likely to result in a response of ‘Probably yes’ rather than ‘No information’. Alternatively, if other (contemporary) trials by the same investigator team have clearly used non-random sequences, it might be reasonable to assume that the current study was done using similar methods."

Passage(s):
<retrieved_paragraph_1>
...
<retrieved_paragraph_k>

\end{spverbatim}
\normalsize

\section{Further Commentary on Inter-Rater Reliability Analysis}
\label{app:inter-rater-reliability-analysis}
Two independent reviewers independently assessed risk of bias for 20 trials using the revised Cochrane ROB2 tool. These assessments were used to compute IAA as reported in the main paper.
Following these annotations, the two reviewers conducted a consensus meeting to better understand discrepancies arising in their annotations. The discussion process involved revisiting the Cochrane Handbook and the official ROB2 guidance document \citep{higgins2011cochrane, Sterne2019RoB2A} to ensure alignment with recommended best practices.

Notable discrepancies emerged in this meeting, classified into four main categories: disagreement at the signaling question level, disagreement at the domain level, differences in judgments between Yes and Probably Yes, and No and Probably No. One reviewer tended to adopt a more conservative approach and tended to opt for ``some concerns'' or ``high risk'' judgments whereas the other reviewer more frequently opted for ``low risk'' ratings when the available information appeared sufficient. This divergence was typical of what has been described in prior research on ROB2, which has noted that even experienced reviewers may differ in how they interpret the level of concern warranted by ambiguous or incomplete reporting \cite{Minozzi2020TheRC}.

Following this consensus meeting, the reviewers were able to reach full agreement across all signaling questions and domains. This calibration process is useful for achieving subsequent consistent application of the ROB2 assessment tool.

\section{Few-Shot Prompting Results}
\label{app:fs-prompting-results}

Results from few-shot prompting experiments are shown in Table~\ref{apptab:fs-results}. The few-shot prompt is created by sampling one example for each class from the gold label annotations, using the same prompt template in App~\ref{app:prompt}. The oracle paragraph is provided for each example, the elaboration for the signaling questions is removed (due to token constraints), and the answer is appended to the end of the prompt in the form \texttt{Answer:<Label>}. Few shot examples sampled are removed from the evaluation set for models.  No substantial differences are observed between the zero- and few-shot settings when models are provided the same number of context passages.
\begin{table*}[tb!]
    \centering
    \small
    \begin{tabular}{lllcccccc}
        \toprule
         Model & Retrieval & D1 & D2 & D3 & D4 & D5 & Micro-avg & Macro-avg \\
        \midrule
        $n$-total & - & 750 & 1278 & 598 & 1027 & 750 & - & - \\
        \midrule
        \multicolumn{9}{l}{\textbf{Zero-shot setting}} \\[0.5mm]
        Llama-3.3-70B-Instruct & k=1 & 0.83 & 0.61 & 0.42 & 0.80 & 0.38 & 0.49 & 0.61 \\
        GPT-3.5-Turbo & k=1 & 0.81 & 0.73 & 0.69 & 0.74 & 0.66 & 0.58 & 0.73 \\
        GPT-4o & k=1 & 0.68 & 0.65 & 0.58 & 0.81 & 0.75 & 0.55 & 0.69 \\
        Claude 3.5-Sonnet & k=1 & 0.69 & 0.68 & 0.66 & 0.87 & 0.76 & 0.60 & 0.73 \\
        \midrule
        \multicolumn{9}{l}{\textbf{Few-shot setting}} \\[0.5mm]
        
        Llama-3.3-70B-Instruct (FS) & k=1 & 0.83 & 0.61 & 0.43 & 0.76 & 0.38 & 0.49 & 0.61 \\
        GPT-3.5-Turbo (FS) & k=1 & 0.81 & 0.70 & 0.65 & 0.80 & 0.71 & 0.60 & 0.73 \\
        GPT-4o (FS) & k=1 & 0.70 & 0.64 & 0.60 & 0.83 & 0.70 & 0.55 & 0.69 \\
        Claude 3.5-Sonnet (FS) & k=1 & 0.69 & 0.68 & 0.65 & 0.87 & 0.77 & 0.60 & 0.73 \\
        \bottomrule
    \end{tabular}
    \caption{Micro-averaged domain-level F1 along with micro- and macro-averaged F1 across signaling questions for few shot retrieval. Zero-shot k=1 results from Table~\ref{tab:results} are reproduced here for reference.
    }
    \label{apptab:fs-results}
\end{table*}

\begin{figure*}[t!]
    \centering
    \includegraphics[width=\textwidth]{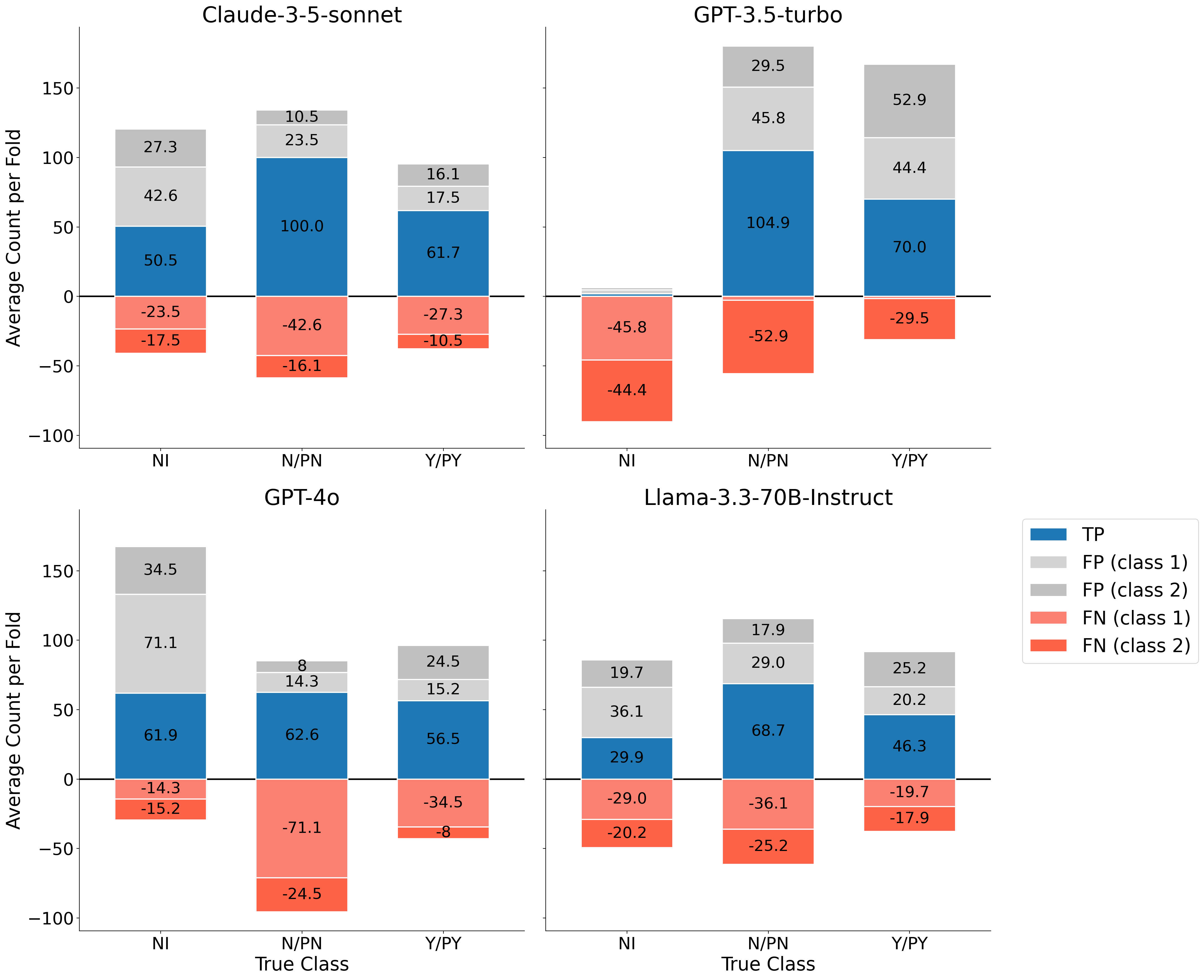}
    \caption{Stacked bar chart showcasing the aggregate true positive (TP) classifications versus false positive/negative (FP/FN) errors made by each model. FPs and FNs are each broken down into two classes, where class 1 (lighter color) are milder errors than class 2 (darker color) (e.g., misclassifying NI and N/PN or Y/PY is less severe than misclassifying N/PN as Y/PY or vice versa). Counts less than 3 have their numbers hidden for chart readability, and full counts are available in Table~\ref{tab:model-comparison}.
    }
    \label{fig:model_comparison}
\end{figure*}

\section{LLM Error Analysis}
\label{app:llm-error-analysis}

\begin{table*}[htbp]
\centering
\small
\begin{tabular}{llrrrrr}
\toprule
Model & Class & True Positives & FP (class 1) & FP (class 2) & FN (class 1) & FN (class 2) \\

\midrule
Llama-3.3-70B-Instruct & NI & 29.9 & 36.1 & 19.7 & 29.0 & 20.2 \\
 & N/PN & 68.7 & 29.0 & 17.9 & 36.1 & 25.2 \\
 & Y/PY & 46.3 & 20.2 & 25.2 & 19.7 & 17.9 \\
\midrule
GPT-3.5-turbo & NI & 1.8 & 2.8 & 1.6 & 45.8 & 44.4 \\
 & N/PN & 104.9 & 45.8 & 29.5 & 2.8 & 52.9 \\
 & Y/PY & 70.0 & 44.4 & 52.9 & 1.6 & 29.5 \\
\midrule
GPT-4o & NI & 61.9 & 71.1 & 34.5 & 14.3 & 15.2 \\
 & N/PN & 62.6 & 14.3 & 8.3 & 71.1 & 24.5 \\
 & Y/PY & 56.5 & 15.2 & 24.5 & 34.5 & 8.3 \\
\midrule
Claude-3.5-Sonnet & NI & 50.5 & 42.6 & 27.3 & 23.5 & 17.5 \\
 & N/PN & 100.0 & 23.5 & 10.5 & 42.6 & 16.1 \\
 & Y/PY & 61.7 & 17.5 & 16.1 & 27.3 & 10.5 \\
\bottomrule
\end{tabular}
\caption{Confusion-matrix summary for the LLMs when answering ROB2 signaling questions.
For each response category Yes/Probably Yes (Y/PY), No/Probably No (N/PN), and No Information (NI), we list the number of true positives (TP) and the false positive (FP) and false negative (FN) counts accrued against the two alternative classes (“class 1” and “class 2”). Higher TP and lower FP/FN values reflect better agreement with the gold label. All values are normalized averages across run configurations.}
\label{tab:model-comparison}
\end{table*}

The 4 LLMs we evaluated exhibit different patterns of answers, though the evaluation metrics are comparable. Performance across models according to micro- and macro-averaged F1 across domains suggests similar performance, qualitative performance is very different between models. We plot error counts in Figure~\ref{fig:model_comparison}, showing true positives (TP), alongside each of the two types of false positives (FPs) and false negatives (FNs). Here, class 1 FP/FN errors are those considered to be less severe (e.g., Y/PY wrongly classified as NI is not as severe as Y/PY wrongly classified as N/PN). We also provide raw counts of these errors in Table~\ref{tab:model-comparison}.

As seen in the figure, GPT-4o is more likely than other LLMs to answer ``No Information'' (large number of FP in the first column) or ``No/Probably No''. Llama-3.3-70B-Instruct, Claude-3.5-Sonnet, and GPT-4o most often predicted ``No Information'' for signaling questions where the true label was No/Probably No. On the other hand, GPT-3.5-Turbo almost never abstains with a ``No Information'' prediction, leading to more false positive errors for the N/PN and Y/PY classes. Claude 3.5-Sonnet was the best performing model evaluated and has fairly comparable false positive rates across ``No/Probably No'' and ``Yes/Probably Yes''. Llama-3.3-70b-Instruct demonstrates a similar answer distribution to Claude 3.5-Sonnet, but with a consistent false positive rate across all 3 classes, but higher false negatives, with ``No/Probably No'' being the highest.

These observed behaviors for over-predicting ``No Information'' or ``No/Probably No'' could stem from safety mechanisms learned during model post-training, which might explain GPT-3.5-Turbo's extreme bias towards ``No/Probably No'' and ``Yes/Probably Yes''. Some domains have questions phrased in a way that requires interpreting numerical data (D2 \& D3) or understanding current best practices in the field (D5); stronger models tend to abstain in these cases and select ``No Information''.
However, in the context of ROB2 assessments, these cautious predictions lead to over-conservative domain- and paper-level labels (high risk judgments) for LLM-supported assessments.

\section{\system Usage Statistics}
\label{app:annotator-para-pref}

Acceptance rates of model answers and rationales versus user-corrected rates are provided in Table~\ref{tab:usage-results-table}. 
Our annotation interface allows users to rate the 3 retrieved paragraphs with a good/bad rating and/or add their own paragraphs from the paper as context. Feedback statistics for retrieved evidence paragraphs are given in Table~\ref{tab:feedback-scores-table}.

\begin{table*}[t!]
    \centering
    \small
    \begin{tabular}{c c c c c}
        \toprule
         & \multicolumn{2}{c}{\textbf{Predictions}} & \multicolumn{2}{c}{\textbf{Rationales}} \\
        Domain & Model (\%) & Expert (\%) & Model (\%) & Expert (\%) \\
        \midrule\\[-0.15in]
        1 & 377 (49.2\%) & 390 (50.8\%) & 430 (56.1\%) & 337 (43.9\%) \\
        2 & 853 (59.2\%) & 588 (40.8\%) & 1117 (77.5\%) & 324 (22.5\%) \\
        3 & 432 (65.2\%) & 231 (34.8\%) & 494 (74.5\%) & 169 (25.5\%) \\
        4 & 591 (64.6\%) & 325 (35.4\%) & 717 (78.3\%) & 199 (21.7\%) \\
        5 & 368 (48.2\%) & 396 (51.8\%) & 485 (63.5\%) & 279 (36.5\%) \\
        \midrule
        Total & 2621 (57.6\%) & 1930 (42.4\%) & 3243 (71.3\%) & 1308 (28.7\%) \\
        \bottomrule
    \end{tabular}
    \caption{Counts and percentages of model-originated versus expert-corrected predictions and explanations across domains for ROB assessments completed using \system.}
    \label{tab:usage-results-table}
\end{table*}

\begin{table*}[t!]
    \centering
    \small
    \begin{tabular}{cccc}
        \toprule
        Domain & Downvotes & Upvotes & User Added Paragraphs \\
        \midrule\\[-0.15in]
        1 & 43 & 207  & 74 \\
        2 & 40 & 120  & 84 \\
        3 & 12 & 48  & 24 \\
        4 & 22 & 64  & 112 \\
        5 & 18 & 41  & 62 \\
        \midrule
        Total & 135 & 480 & 356  \\
        \bottomrule
    \end{tabular}
    \caption{Feedback on evidence passages provided by users by domain at the question level, for the \system subset, i.e., each number corresponds to the number of questions in that domain for which a user provided a downvote, an upvote, or added paragraph, as opposed to the total number of downvotes or upvotes etc.}
    \label{tab:feedback-scores-table}
\end{table*}

\end{document}